\pdfoutput=1
\documentclass{article}

     \PassOptionsToPackage{numbers, compress}{natbib}
     \usepackage[preprint]{neurips_2019}

\usepackage[utf8]{inputenc} % allow utf-8 input
\usepackage[T1]{fontenc}    % use 8-bit T1 fonts
\usepackage{hyperref}       % hyperlinks
\usepackage{url}            % simple URL typesetting
\usepackage{booktabs}       % professional-quality tables
\usepackage{amsfonts}       % blackboard math symbols
\usepackage{nicefrac}       % compact symbols for 1/2, etc.
\usepackage{microtype}      % microtypography
\usepackage{verbatim}
\usepackage{graphicx}
\usepackage{mathtools}

\DeclarePairedDelimiter{\floor}{\lfloor}{\rfloor}
\usepackage{amsmath}
\usepackage{wrapfig}
\usepackage[flushleft]{threeparttable}
\newcommand{\bd}[1]{\textbf{#1}}

\title{Revisiting Feature Alignment for One-stage Object Detection}

\author{%
  Yuntao Chen$^{1,3}$ \quad Chenxia Han \quad Naiyan Wang$^{2}$  \quad Zhaoxiang Zhang$^{1,3,4}$\\
  $^{1}$ University of Chinese Academy of Sciences \qquad
$^{2}$ TuSimple\\
$^{3}$ Center for Research on Intelligent Perception and Computing, CASIA\\
$^{4}$Center for Excellence in Brain Science and Intelligence Technology, CAS\\
{\tt\small \{chenyuntao2016, zhaoxiang.zhang\}@ia.ac.cn }\\
{\tt\small \{chenxiahan18, winsty\}@gmail.com} \\
}

\begin{document}

\maketitle

\begin{abstract}
Recently, one-stage object detectors gain much attention due to their simplicity in practice.
Its fully convolutional nature greatly reduces the difficulty of training and deployment compared with two-stage detectors which require NMS and sorting for the proposal stage.
However, a fundamental issue lies in all one-stage detectors is the misalignment between anchor boxes and convolutional features, which significantly hinders the performance of one-stage detectors.
In this work, we first reveal the deep connection between the widely used \verb|im2col| operator and the RoIAlign operator. Guided by this illuminating observation, we propose a RoIConv operator which aligns the features and its corresponding anchors in one-stage detection in a principled way. 
We then design a fully convolutional AlignDet architecture which combines the flexibility of learned anchors and the preciseness of aligned features. 
Specifically, our AlignDet achieves a state-of-the-art mAP of 44.1 on the COCO \emph{test-dev} with ResNeXt-101 backbone.
\end{abstract}

\section{Introduction}
Object detection is a location sensitive task. A location sensitive task requires location-sensitive features, which means the input feature should vary with its reference box. Thus aligning the features with its corresponding bounding box is at the core of object detection. From the early sliding window methods to the state-of-the-art generalized R-CNNs~\cite{faster-rcnn, mask-rcnn}, detectors make efforts to ensure the feature alignment. The alignment of feature and box can be done in the image-level like R-CNN~\cite{rcnn}, in the feature-level like Fast R-CNN~\cite{fast-rcnn}, and even in the result-level like R-FCN~\cite{rfcn}. The alignment can be achieved by crop and resize, or various kinds of region feature extractors such as RoIPooling~\cite{fast-rcnn} and RoIAlign~\cite{mask-rcnn}.

The introduction of the anchor box~\cite{faster-rcnn} changes the game. Anchor boxes are a set of virtual rectangles of different scales and aspect ratios serving as references for classification and regression. Anchor shapes are hand-crafted, chosen by clustering ground-truth bounding boxes or learned from previous stages in the network~\cite{refinedet, guided-anchor}. Each spatial location of a feature map is associated with multiple anchors of different shapes. \emph{Sharing feature for multiple anchor boxes violate the one-to-one correspondence between reference boxes and features, which breaks the strict location sensitivity of object detection.}   % Anchor boxes can be of different shapes and Since anchors are virtual, the features bounded by anchor boxes are not the same features associated with the anchor boxes. In fact, the misalignment is inherent as the same feature is always mapped to multiple different anchors.

One-stage detectors~\cite{ssd,retinanet,yolov3} suffer most from the misalignment due to the lack of explicit alignment operations like RoIPooling in two-stage detectors. The multi-scale features widely adopted in one-stage detectors partially resolve this misalignment by assigning anchors of different scales to proper feature levels. Nevertheless, the implicit alignment conducted by multi-scale features only relieves the misalignment incurred by scales. They can not help the aspect ratios induced misalignment. Besides, feature adaptation~\cite{refinedet,guided-anchor} can also be seen as a simple form of feature alignment.  

In this work, by discovering the deep connection between \verb|im2col|~\cite{im2col} and RoIAlign, we propose a novel RoI Convolution (RoIConv) which performs accurate feature alignment in one-stage detectors for the first time. Our RoIConv shares the same computation complexity the vanilla convolution and can be seamlessly integrated into any existing one-stage detectors in a plug-and-play manner. Based on that, we also propose a Fully Convolutional AlignDet detector which perfectly combines the flexibility of learned anchors and the preciseness of our RoIConv. \emph{Our method enjoys the explicit alignment of two-stage detectors while remains the fully convolutional nature and computation cost of one-stage detectors, getting the bests of both worlds.}

To summarize, our contributions are as following:

\begin{itemize}
	\item We propose a plug-and-play RoI Convolution (RoIConv) operation, which performs exact feature alignment densely for one-stage detectors for the first time.
	\item We propose a Fully Convolutional AlignDet model which fully utilizes the benefits of learned anchors with our RoIConv.
	\item Our methods achieve state-of-the-art 44.1 mAP with the ResNeXt-101 FPN backbone, improving over the strong RetinaNet baseline by 3.3 mAP while keeping the same speed.
\end{itemize}

\section{Related Works}
\paragraph{One-stage Detectors}
Redmon \emph{et al.} propose YOLO~\cite{yolo}, which is first work to use an end-to-end convolution network for detection directly on the dense feature map.
SSD~\cite{ssd} introduces anchors and multi-scale feature maps into one-stage detection.
RetinaNet~\cite{retinanet} proposes focal loss to address the overwhelming easy background samples introduced by dense multi-scale feature maps.
Keypoint-based detection methods~\cite{cornernet, centernet, centernet2, extremenet} have shown promising results by detecting and grouping the corners for objects.
Besides, recently emerging anchor-free methods~\cite{fsaf, foveabox, fcos, densebox, meta-anchor} explore the potential of detecting objects without virtual anchor boxes.

\paragraph{Feature Alignment in Object Detection}
SPP-net~\cite{spp} is the first work to extract fixed-length features from candidate windows in convolution networks.
RoIPooling~\cite{fast-rcnn} improves over SPP-net by enabling end-to-end training.
Both SPP-net and RoIPooling round the sub-region to the nearest integer boundary which incurs quantization errors.
To address the quantization error of RoIPooling, RoIAlign~\cite{mask-rcnn} uses bilinear interpolation to compute the exact values at sampled locations in each RoI bins, showing significant gains for localization.
Deformable RoIPooling~\cite{deformable} adds offsets to each sub-region for RoIPool, bringing adaptiveness for the region feature.
Guided Anchor~\cite{guided-anchor} tries to adapt features for learned anchors with anchor-guided deformable convolutions.

\paragraph{Cascaded Object Detection System}
RefineDet~\cite{refinedet} introduces an anchor refinement module to adjust center points and sizes of anchors, providing better reference boxes for further regression.
Cai \emph{et al.} propose the Cascade R-CNN~\cite{cascade}, which improves the quality of proposals in cascaded stages with increasing IoU thresholds. Region features are re-extracted for the refined proposals in each stage.

\section{Pilot Experiement}
\subsection{Multi-scale Features and One-stage Object Detection}
\label{sec:pilot}
Along with the development of detectors, multi-scale feature maps~\cite{ssd,fpn} play a central role for the handling scale variations of objects.
We argue that multi-scale feature maps are especially essential for one stage detectors since they lack the ability to align features and corresponding bounding boxes.
To demonstrate the importance of multi-scale features, let us consider two settings for one-stage detectors.
For a RetinaNet~\cite{retinanet} detector with a $P_3-P_7$ FPN backbone, the strides for different pyramid levels are $\{8, 16, 32, 64, 128\}$. 
When equipped with an anchor box of a scale factor of 4, this detector yields a set of anchor boxes of size $\{32^2, 64^2, 128^2, 256^2, 512^2\}$ across five scales of features.
To demonstrate the effectiveness of multi-scale features, we can construct a detector with only the $P_4$ feature, which has a single stride of 16. 
Given a set of anchors with scale factors of $\{2, 4, 8, 16, 32\}$, the second detector yields the same set of anchor boxes, but on a single scale of feature.
\emph{However, the former detector gives an mAP of 32.4 on the COCO minival set, but the later one only gives an mAP of 20.4.}
We also test the same setting for a standard two-stage detector Faster R-CNN. It only shows a minor drop (mAP 33.9 down to 31.6) when using only one scale of features. The drastic performance drop for the one-stage detector is unusual, considering two-stage detectors that adopt a single-scale feature map can still achieve comparable results.

\begin{table}[hbpt]
\caption{Comparing performance drop of Faster R-CNN and RetinaNet without FPN on the COCO \textit{minival} set. Two-stage detectors achieve far better results than one-stage ones without multi-scale features.}
\label{tb:t1-pilot}
\small
\renewcommand\arraystretch{1.2}
\centering
\begin{tabular}{ccccccccc}
	\toprule
	Model        & Backbone        & Feature                   & AP     & AP$_{50}$ & AP$_{s}$ & AP$_{m}$ & AP$_{l}$ & $\Delta$AP \\
	\midrule
	Faster R-CNN~\cite{fpn} & ResNet-50 FPN   & $\{P_2 - P_5\}$  & 33.9	& 56.9	& 17.8 & 37.7 & 45.8 & - \\
	Faster R-CNN~\cite{fpn} & ResNet-50       & $C_4$                     & 31.6	& 53.1	& 13.2 & 35.6 & 47.1 & $2.3\downarrow$ \\
	RetinaNet    & ResNet-50 FPN   & $\{P_3 - P_7\}$  & 32.4	& 52.9	& 17.5 & 35.9 & 43.0 & - \\
	RetinaNet    & ResNet-50       & $P_4$                     & 20.4	& 37.4	& 7.3  & 24.6 & 32.3 & $12.0\downarrow$ \\
	\bottomrule
\end{tabular}
\end{table}

A prominent difference between one-stage and two-stage detectors is that one-stage detectors lack RoI feature extractors like RoIPooling~\cite{fast-rcnn} and RoIAlign~\cite{mask-rcnn}. 
RoI feature extractor provides aligned features for each RoI. But for one-stage detectors, all anchor boxes(one-stage counterpart of RoI in two-stage detectors) share the same features for the same spatial location. Multi-scale feature maps merely alleviate this misalignment issue by limiting the scale of each feature map.
We hypothesize that \emph{the misalignment of features and anchor boxes leads to the catastrophic performance degradation}.

\section{Methods}

\subsection{Im2Col is a special RoIAlign}
\begin{figure}[h]
	\centering
	\makebox[\textwidth]{\includegraphics[trim={0 7cm 0 0}, clip, width=0.9\textwidth]{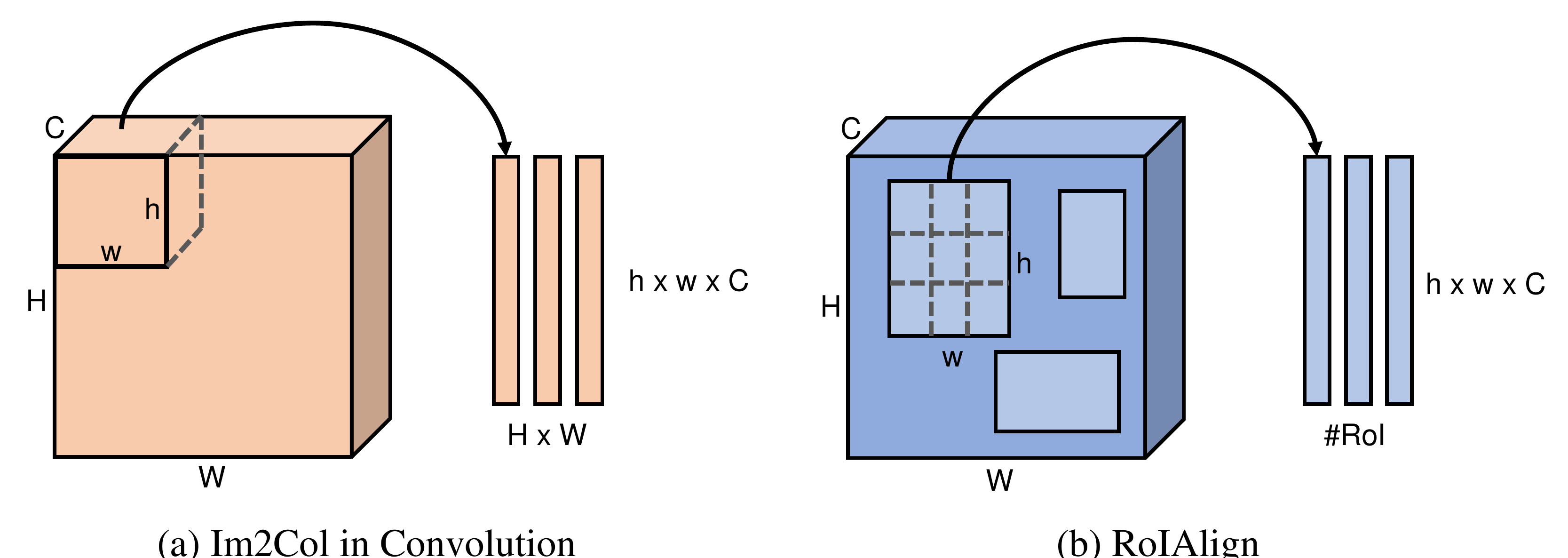}}
	\caption{A demonstration of the resemblance between im2col and RoIAlign.}
	\label{fig:f1-im2col}
\end{figure}

To fully reveal the essence of feature misalignment, we now take a deeper look at RoI feature extractors of two-stage detectors.
Taking RoIAlign~\cite{mask-rcnn} as an example, it first divides an RoI evenly into $h \times w$ sub-regions and then take the center point\footnote{This is only true for sampling ratio = 1. Since the sampling ratio has little impact as indicated in \cite{fpn}, we stick to sampling ratio = 1 to simplify the discussion.} of each sub-region on the bi-linear interpolated feature map.
The features extracted from each sub-region are then concatenated to give a feature  $f \in \mathbb{R} ^ {h \times w \times C_{in}}$ for the given RoI.
As shown in Figure~\ref{fig:f1-im2col}, the whole process strikes a remarkable resemblance of the \verb|im2col|~\cite{im2col} operation, which is a core part in the implementation of convolution. \verb|Im2col| transforms the 3-D input feature tensor $F \in \mathbb{R} ^ {C_{in} \times H \times W}$ into a 2-D tensor $\tilde{F} \in \mathbb{R} ^ {hwC_{in} \times HW}$, where $h$ and $w$ are the height and width of the convolution kernel. Each column in $\tilde{F}$ represents a tile which the convolution kernel slides on.
The only difference is \verb|im2col| operates on a fixed set of spatial locations on the input feature map, whereas RoIAlign operates on locations defined by the RoI. \verb|Im2col| is essentially a special case of RoIAlign. Since convolution is the combination of \verb|im2col| and fully connected layer, \emph{one-stage detectors are implicitly performing RoIAlign on the backbone feature maps with fixed bounding box size}.

After revealing the connection between RoIAlign and \verb|im2col|, it is natural to find out what is the RoI of a convolution.
For a $h \times w$ convolution, its sampling locations for $(X, Y)$ of the output feature map are given as: 
\begin{equation}
		Loc_{\text{conv}}(i, j) = \left( X - \floor{\frac{h}{2}}  + i + 0.5, Y - \floor{\frac{w}{2}} + j + 0.5 \right),
\end{equation}
and for a RoIAlign with a RoI of $(x_1, y_1, x_2, y_2)$ and a output of $h \times w$, its sampling locations on the feature map of stride $S$ are given as 
\begin{equation}
    Loc_{\text{RoI}}(i, j) = \left( \dfrac{hx_1 + (x_2 - x_1)(i + 0.5)}{hS}, \dfrac{wy_1 + (y_2 - y_1)(j + 0.5)}{wS} \right),
\end{equation}
where $i \in \{0, 1, \dots, h-1\}$ and $j \in \{0, 1, \dots, w-1\}$ for both operators.
Solving $Loc_{\text{conv}} = Loc_{\text{RoI}}$ gives 
\begin{equation}
\label{eq:e3-conv-roi}
\begin{aligned}
(x_1, y_1) &= \left((X - \floor{\frac{h}{2}})S , (Y - \floor{\frac{w}{2}})S\right), \\
(x_2, y_2) &= \left((X - \floor{\frac{h}{2}} + h)S, (Y - \floor{\frac{w}{2}} + w)S\right).
\end{aligned}
\end{equation}

Equation~\ref{eq:e3-conv-roi} shows that a $h \times w$ convolution on a feature map of stride $S$ is equivalent to a $hS \times wS$ RoIAlign for each input location, followed by a fully connected layer with weight $W \in \mathbb{R} ^ {C_{out} \times (h \times w \times C_{in})}$.

We now revisit the example introduced in the pilot experiment. 
For the $P_4$ feature map of the FPN, the total stride is 16, and thus a $3 \times 3$ convolution on $P_4$ feature gives an RoI of $48 \times 48$. 
This RoI partially aligns with the anchor boxes of size $32^2$ and $64^2$, but incurs heavy misalignment for the other three anchors. 
Stacking multiple convolutions in the detection head may increase the implicit RoI range for each spatial location, but the misalignment between a single RoI and multiple anchors persist. 
Detectors like SSD and RetinaNet utilize multi-scale feature maps to solves this problem. 
By assigning anchors to feature maps of the proper stride, the RoI of the convolution and the anchor matches. 
We take RetinaNet as an example. 
The implicit RoIs of convolutions for the $P_3 - P_7$ FPN spans from $24^2$ to $384^2$, covering the anchors from $32^2$ to $512^2$. This operation partially addresses the misalignment to some degree, but it still cannot handle harder cases such as extreme aspect ratio, etc.
\subsection{RoI Convolution}
%Although the multi-scale feature maps could relieve the misalignment incurred by anchor scales, but there are other types of anchor/feature misalignment such as those induced by anchors of different aspect ratios and learned anchors.
To address the challenges mentioned above, we devise an operator called RoIConv, which aligns features and corresponding anchor boxes in a principled way.
By inspecting Equation~\ref{eq:e3-conv-roi} closely, we can find that the misalignment between the feature and the anchor box is indeed caused by the misalignment between the implicit RoI of the convolution and the actual bounding box. 
Inspired by deformable convolution\cite{deformable}, we can adaptively sample locations of the convolution by introducing an offset map for each location. Deformable convolution learns a $2 \times h \times w$ for each location of the output feature map. Each pair of offset $(O_x, O_y)$ describe the deviation from the regular convolution sampling points.
Different from deformable convolution, our offsets are now calculated as the difference of the pre-defined anchor box and the implicit RoI instead of learned.
For a $h \times w$ RoIConv on a feature map of stride $S$ and its corresponding anchor box $(x_1, y_1, x_2, y_2)$, the offsets $O$ for a specific location $(X, Y)$ on the output feature map are given as
\begin{equation}
\label{eq:roiconv}
\begin{split}
O_x(i) = Loc_{\text{anchor}}(i) - Loc_{\text{conv}}(i) = \dfrac{x_1}{S} - X + \floor{\frac{h}{2}} + \left( \dfrac{x_2 - x_1}{hS} - 1 \right)(i + 0.5)&, \\
O_y(j) = Loc_{\text{anchor}}(j) - Loc_{\text{conv}}(j) = \dfrac{y_1}{S} - Y + \floor{\frac{w}{2}} + \left( \dfrac{y_2 - y_1}{wS} - 1 \right)(j + 0.5)&, \\
\forall (i, j) \in \{0, 1, \dots, h-1\} \times \{0, 1, \dots, w-1\}&.
\end{split}
\end{equation}
It is worth noting that our RoIConv requires no addition computation compared with the vanilla convolution, which helps its seamless integration into any existing one-stage detectors.
The offsets are a linear combination of $(x_1, y_1, x_2, y_2)$ and $(X, Y)$, which can be obtained with a $1 \times 1$ convolution and an element-wise addition. This help to keep one-stage detectors fully convolutional.

\subsection{Fully Convolutional AlignDet}
\label{sec:s4-aligndet}
\begin{figure}[t]
	\centering
	\makebox[\textwidth]{\includegraphics[trim={0 9cm 0 0}, clip, width=\textwidth]{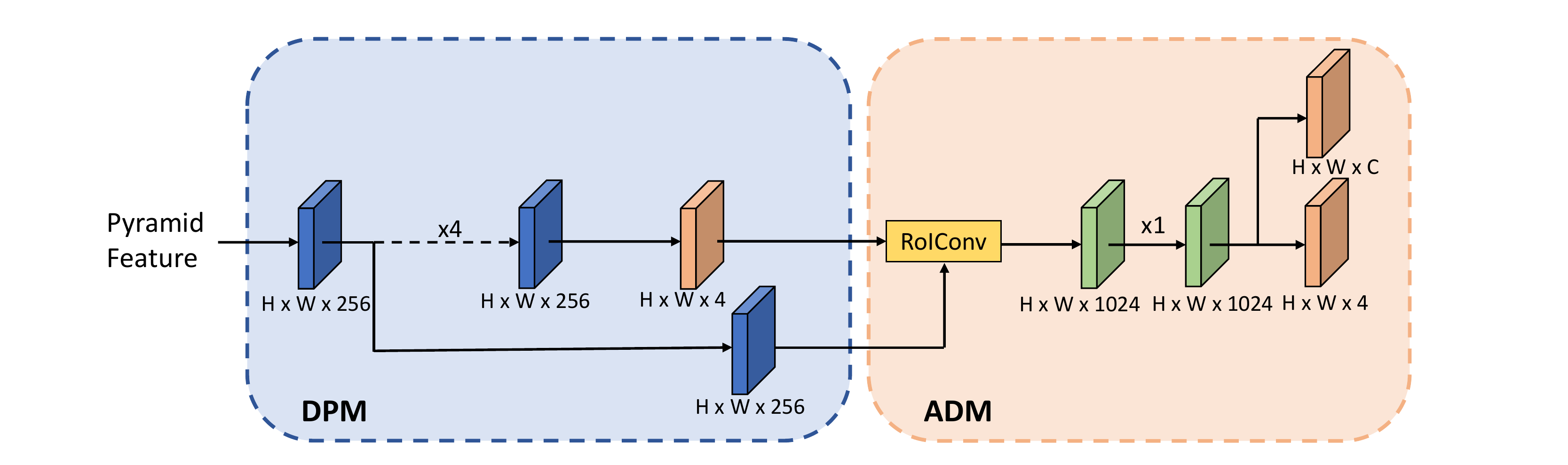}}
	\caption{The architecture of the proposed Fully Convolutional AlignDet during inference. The DPM stands for our dense proposal module which is responsible for learning adapative anchors for the ADM. The ADM stands for our aligned detection module which first aligns the backbone feature with the learned anchor. The aligned features are fed into the dense detection head for the final classification and bounding box regression.}
	\label{fig:f2-arch}
\end{figure}

With the proposed RoI Convolution, we can explore more flexible anchor settings. 
Following previous learned anchor works~\cite{refinedet, guided-anchor}, we propose Fully Convolutional AlignDet, which consists of a dense proposal module(DPM) and an aligned detection module(ADM).
The dense proposal module could be any network that gives a dense bounding box prediction on the feature maps, including RPN~\cite{faster-rcnn}, SSD~\cite{ssd}, RetinaNet~\cite{retinanet} and even recently proposed anchor-free detectors~\cite{fsaf, foveabox, fcos, densebox, meta-anchor}. The dense proposal module learns the bounding box distribution from data and thus liberates us from setting anchor manually.
As shown in Figure~\ref{fig:f2-arch}, the aligned detection module consists of a RoIConv for aligning the backbone feature with the learned anchors and the subsequent detection head for predicting the final scores and bounding boxes. Feature alignment is especially important for that learned anchors are far more varying in scales, aspect ratios, and locations than hand-crafted ones. 

\paragraph{Comparison with Other Feature Alignment Alternatives}
There are also other dense detectors adopting the learned anchor paradigm which requires feature alignment. RefineDet~\cite{refinedet} consists of an anchor refinement module which refines the initial anchor boxes and an object detection module which predicts the final class and bounding box of the refined anchors. RefineDet performs feature adaptation by a vanilla $3 \times 3$ convolution, which serves as a baseline for feature alignment. Guided Anchor is an anchor-free dense detector which directly predicts the shape of the anchor box for each spatial location. Due to the varying shape of the learned anchor, Guided Anchor~\cite{guided-anchor} tries to adapt backbone features to fit the anchor shape by learning deformable offsets from the predicted anchor shapes. The feature adaptation in Guided Anchor is less precise compared with our RoIConv in two ways. First, it only considers the shape for anchors but ignores the location of anchor boxes. As indicated in Equation~\ref{eq:roiconv}, the optimal offsets comprise both the shape and the location of the anchor box. Second, the learned offsets are just heuristic approximation of the offsets calculated from Equation~\ref{eq:roiconv}. Our method can mathematically guarantee the strict alignment between features and their corresponding anchors, while theirs cannot.
\begin{figure}[th]
	\centering
	\makebox[\textwidth]{\includegraphics[trim={0 11cm 0 0}, clip, width=\textwidth]{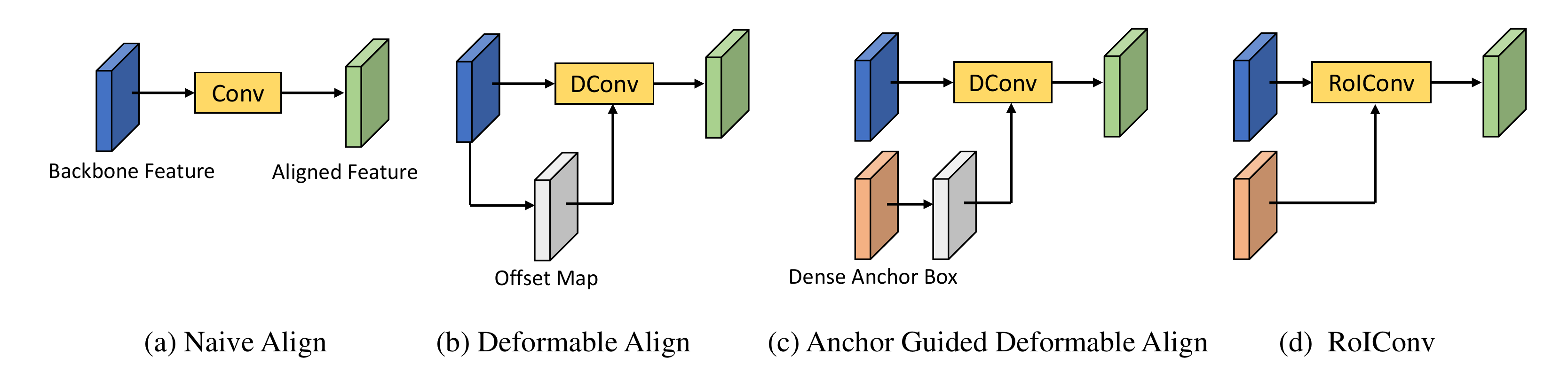}}
	\caption{A comparison between different feature align methods. (a) is the approach of RefineDet which aligns feature by vanilla convolution; (b) tries to align feature by learning offsets for each locations, the supervision signal comes from final loss; (c) is the approach of Guided Anchor which improves overs (b) by leveraging anchor shape information to generate offsets; (d) is the proposed RoIConv which directly encodes the anchor box as offsets to perfectly align features and anchors. From (a) to (d) the feature alignment is more and more precise.}
	\label{fig:f3-compare}
\end{figure}

\section{Experiments}
\subsection{Implementation Details}
We train the models on the COCO~\cite{coco} \emph{trainval35k} split and report results in mAP on the COCO \emph{minival} split.
We use RetinaNet~\cite{retinanet} with a single anchor of scale $4$ and aspect ratio $1:1$ as our DPM, followed by the ADM.
DPM only does bounding regression during the test phase. 
We set loss weight of both DPM and ADM to $1.0$. The focal losses in both DPM and ADM are set to $\alpha=0.25$ and $\gamma = 2$.
Specifically, we use ResNet-50 FPN~\cite{fpn} and ResNet-101 FPN as our backbone. We use feature pyramids from $P_3$ to $P_7$.
The backbone is initialized from ImageNet~\cite{imagenet} 1k pre-training, and the newly added FPN layers are initialized with He initialization~\cite{he2015delving}. The newly added head layers are initialized with Gaussian initializer with $\sigma=0.01$. We freeze the backbone up to $conv_2$ as well as all BN parameters. 
Input images are resized to a short side of 800 and a long side not exceeding 1333 and horizontal flip is adopted during training.
We train all models in SGD for 90k iterations with a starting learning rate of 0.01 and divide the learning rate by 10 in 60k and 80k iterations using a total batch size of 16 over 8 GPUs. We adopt learning rate warmup for 500 iterations. The weight decay is set to 0.0001.
NMS with IoU threshold 0.5 is adopted for post-processing.
\subsection{RoI Convolution for Single-scale One-stage Detection}
We first demonstrate the importance of feature alignment by comparing the performances of RetinaNets based on a single-scale feature map with and without RoIConv. We use the same settings as in the pilot experiments in Sec.~\ref{sec:pilot}. For RetinaNet + RoIConv, we add one $3 \times 3$ RoIConv on $P_4$ with the pre-defined anchors as RoIs. From the results of Table~\ref{tb:t2-roiconv}, we can see that RoIConv is an effective and efficient way for feature alignment. With merely one extra $3 \times 3$ convolution, the highly misaligned single-scale RetinaNet recovers an mAP of 5.0.
\begin{table}[hbpt]
\caption{Comparing single-scale feature one-stage detectors with and without RoIConv on the COCO \textit{minival} set.}
\label{tb:t2-roiconv}
\small
\renewcommand\arraystretch{1.25}
\centering
\begin{tabular}{cccccccc}
	\toprule
	Model        & Backbone          & Feature   & AP     & AP$_{50}$ & AP$_{s}$ & AP$_{m}$ & AP$_{l}$ \\
	\midrule
	RetinaNet    & ResNet-50         & $P_4$     & 20.4	& 37.4	& 7.3  & 24.6 & 32.3  \\
	RetinaNet + RoIConv & ResNet-50  & $P_4$     & 25.4	& 42.2	& 8.6  & 28.6 & 40.9  \\
	\bottomrule
\end{tabular}
\end{table}

\subsection{Fully Convolutional Aligned Detection}
We now present the results of our Fully Convolutional AlignDet. We compare our methods with and without ADM. AlignDet w/o ADM is essentially a RetinaNet with a single anchor. We compare our methods with the original RetinaNet with 3 scales and 3 aspect ratio anchors. 
As shown in Table~\ref{tb:t3-aligndet}, despite that our ADM is simply a 1024c $7 \times 7$ RoIConv followed by a 1024c $1 \times 1$ convolution, AlignDet achieves 5.5/5.3 mAP improvement over the baseline without ADM for ResNet-50/101. The results prove the effectiveness of the feature alignment for one-stage object detection. AlignDet also achieves an improvement of 2.2/2.1 mAP over RetinaNet for ResNet-50/ResNet-101 backbones. Compared with original RetinaNet, AlignDet uses minimal anchors, which liberates the users from cumbersome hyper-parameter selection for anchors. This shows the learned anchors are on par with the expert-crafted anchors, which echos with recent trends in designing anchor-free one-stage detectors.
\begin{table}[hbpt]
\caption{Comparing AlignDet with RetinaNet on the COCO \textit{minival} set.}
\label{tb:t3-aligndet}
\small
\renewcommand\arraystretch{1.25}
\centering
\begin{tabular}{c|c|c|cc|ccc}
	\toprule
	Model              & Backbone        & \#Anchor  & AP     & AP$_{50}$ & AP$_{s}$ & AP$_{m}$ & AP$_{l}$ \\
	\midrule
	RetinaNet          & ResNet-50 FPN   & 9 & 35.7	& 55.0	& 18.9  & 38.9 & 46.3  \\
	RetinaNet          & ResNet-101 FPN  & 9 & 37.8	& 57.5	& 20.2  & 41.1 & 49.2  \\
	AlignDet w/o ADM   & ResNet-50 FPN   & 1 & 32.4	& 52.9	& 17.5  & 35.9 & 43.0  \\
	AlignDet w/o ADM   & ResNet-101 FPN  & 1 & 34.5	& 55.7	& 18.1  & 38.4 & 45.6  \\
	AlignDet           & ResNet-50 FPN   & 1 & 37.9	& 57.7	& 21.5  & 41.1 & 50.8  \\
	AlignDet           & ResNet-101 FPN  & 1 & 39.8	& 60.0	& 22.6  & 43.4 & 52.8  \\
	\bottomrule
\end{tabular}
\end{table}

\subsection{Ablations}
\begin{figure}[ht]
  \centering
  \includegraphics[trim={5pt 5pt 0pt 5pt}, width=0.7\textwidth]{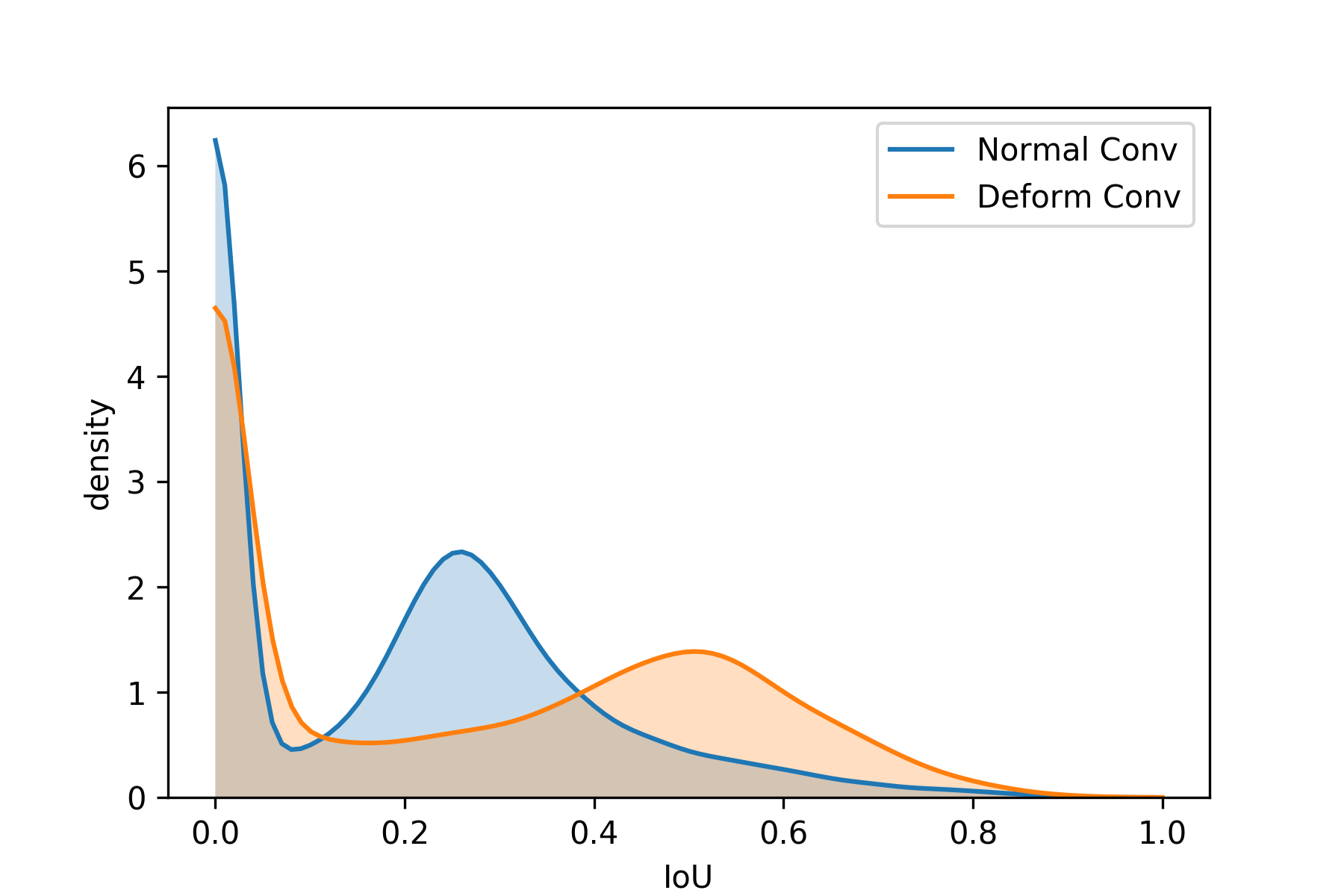}
  \caption{The distribution of IoUs between the implicit RoIs of convs and the refined anchors. The implicit RoI of DeformConv is obtained from the circumscribed rectangle of offset points.}
  \label{iou}
\end{figure}

\paragraph{Variants for Feature Alignment} As discussed in Section~\ref{sec:s4-aligndet}, there are different methods for feature alignment. We conduct controlled experiments in this section to find out the most effective design choice. For all variants, we employ a $3 \times 3$ convolution with an output channel of 256 regardless of the convolution types to ensure the same parameter number. For (b) and (c) the $1 \times 1$ convolutions for the offset generation are initialized with a Gaussian of $\sigma = 0.01$ to ensure a roughly zero offset from the beginning. For (c) we first derive the height and width for each anchor box and then apply Batch Normalization to address the large variance of learned anchor shapes. The normalized anchor shapes are then used to generate offsets. From the results of Table~\ref{tb:t4-variant}, we can see that the proposed RoI Convolution surpasses all other variants for feature alignment. Surprisingly, (b) and (c) do not improve over the vanilla convolution. To better understand this, we decode the implicit RoIs from the offsets learned of deformable convolution and calculate the IoU between the implicit RoIs the learned anchors. As shown in Figure ~\ref{iou}, deformable convolutions actually learn better alignment than vanilla $3\times 3$ convolutions but the overall alignment with learned anchors is still far from ideal. We hypothesize that it may due to the supervision from the classification loss tends to drive the focus of the convolution to the most discriminative part of object, which may hurt the alignment.

\begin{table}[hbpt]
\caption{Comparing different feature alignment variants on the COCO \textit{minival} set. All results are obtained with ResNet-50 FPN. ADM head consists of four $3 \times 3$ convolutions with 256 channels.}
\label{tb:t4-variant}
\small
\renewcommand\arraystretch{1.25}
\centering
\begin{tabular}{c|c|ccc|ccc}
	\toprule
		& 	Variants	     		   & AP   & AP$_{50}$ & AP$_{75}$ & AP$_{s}$ & AP$_{m}$ & AP$_{l}$ \\
	\midrule
	(a)	&	Conv		              & 35.2 & 54.7 & 37.9 & 19.4 & 38.4 & 47.7 \\
	(b)	&	Deform Conv				& 35.3 & 55.0 & 38.0 & 19.0 & 38.7 & 47.1 \\
	(c)	&	Anchor Guided Deform Conv  & 35.2 & 54.8 & 37.9 & 18.5 & 38.7 & 47.3 \\
	(d)	& RoI Conv                   & 36.2 & 56.1 & 39.2 & 19.2 & 39.7 & 48.5 \\
	\bottomrule
\end{tabular}
\end{table}

\paragraph{Label Assignment}
In a cascaded detection pipeline, different parts should be specialized for different purposes. In our Fully Convolutional AlignDet, the DPM is specialized for refining the initial anchor boxes, so we lower the ground truth matching criterion to increase the training samples for the DPM regressor. These thresholds only affect the label assignment during training and all refined anchors are used for prediction during testing. Due to the multi-threshold nature of mAP@0.5:0.95 of COCO, a correct detection box of 0.95 IoU has ten times larger weights of boxes of 0.5 IoU, so we raise the ground truth matching criterion for ADM to bias towards high IoU boxes. 
\begin{table}[hbpt]
\caption{Comparing different label assignment strategies for DPM and ARM. All results are obtained with ResNet-50 FPN on the COCO \textit{minival} set. fg / bg means the IoU thresholds for foreground and background label assignment, respectively. ADM consists of a 256c $3 \times 3$ RoIConv and four 256c $3 \times 3$ convolutions.}
\label{tb:t5-label}
\small
\renewcommand\arraystretch{1.25}
\centering
\begin{tabular}{cc|cccc|ccc}
	\toprule
	DPM fg / bg & ADM fg / bg & AP   & AP$_{50}$ & AP$_{75}$ & AP$_{90}$ & AP$_{s}$ & AP$_{m}$ & AP$_{l}$ \\
	\midrule
	0.5 / 0.4 & 0.5 / 0.4     & 34.2 & 55.8 & 36.3 & 7.5  & 18.5 & 37.4 & 45.7 \\
	0.5 / 0.4 & 0.6 / 0.6     & 35.6 & 55.9 & 38.3 & 9.7  & 19.5 & 38.5 & 47.2 \\
	0.5 / 0.4 & 0.7 / 0.7     & 35.5 & 55.1 & 38.6 & 10.4 & 19.1 & 38.6 & 47.3 \\
	\midrule
	0.4 / 0.3 & 0.6 / 0.6     & 36.0 & 56.6 & 38.5 & 10.2 & 19.5 & 39.2 & 48.2 \\
	0.4 / 0.3 & 0.7 / 0.7     & 36.2 & 56.1 & 39.2 & 11.3 & 19.2 & 39.7 & 48.5 \\
	\bottomrule
\end{tabular}
\end{table}

\paragraph{RoI Convolution Design}
We now explore different design choices of RoIConv. As shown in Table~\ref{tb:t6-roiconv}, the performance improves steadily as the kernel size increases. A large kernel creates dense sampling points which minimize information loss during the process of alignment. High dimension output features also help to reduce information loss. Take a $7 \times 7$ RoIConv as an example. \verb|Im2col| generates a 12544-D feature for each anchor box. Increasing the output channel of RoIConv allows us to preserve more information for the final prediction. A 1024-D output feature with a $1 \times 1$ convolutions as head already outperforms 256-D output feature with four $3 \times 3$ convolutions. Although large kernel size and high dimension output incur substantial computation cost, AlignDet still runs reasonable fast on modern hardware. As shown in the last column of Table~\ref{tb:t6-roiconv}, our $3 \times 3$ variant gains a 1.0 mAP improvement over the original RetinaNet while being 15\% faster with fewer anchors. To further accelerate our AlignDet, we propose the $7 \times 7^\dagger$ variant which reduces the kernel size of RoIConv for the $P_3$ feature to $3 \times 3$. The $7 \times 7^\dagger$ variant is even faster than the original RetinaNet while being 1.9 mAP better.
\begin{table}[hbpt]
\caption{Comparing different RoIConv designs on the COCO \textit{minival} set. All results are obtained with ResNet-50 FPN. Speeds are measured on a NVIDIA 2080 GPU by averaging 100 runs.}
\label{tb:t6-roiconv}
\small
\renewcommand\arraystretch{1.25}
\centering
\begin{tabular}{ccc|ccc|cccc}
	\toprule
	 Kernel Size & Out Channels  & ADM head & AP & AP$_{50}$ & AP$_{75}$ & AP$_{s}$ & AP$_{m}$ & AP$_{l}$ & Speed \\
	\midrule
	$3\times 3$ & 256 & 4 \emph{conv} 256c & 36.2 & 56.1 & 39.2 & 19.2 & 39.7 & 48.5 & 47 ms\\
	$3\times 3$ & 1024 & 1 \emph{conv} 1024c  & 36.7 & 56.9 & 40.4 & 20.1 & 40.4 &  49.2 & 49 ms\\
	$5\times 5$ & 1024 & 1 \emph{conv} 1024c  & 37.2 & 57.1 & 40.7 & 20.6 & 40.3 & 49.9 & 63 ms\\
	$7\times 7$ & 1024 & 1 \emph{conv} 1024c  & 37.9 & 57.7 & 41.7 & 21.5 & 41.1 & 50.8 & 86 ms\\
	$7\times 7^\dagger$ & 1024 & 1 \emph{conv} 1024c & 37.6 & 57.3 & 41.6 & 20.7 & 41.0 & 50.4 & 56 ms\\
	\midrule
	RetinaNet  & - & - & 35.7 & 55.0	 & 38.5 & 18.9  & 38.9 & 46.3 & 58 ms\\ 
	\bottomrule
\end{tabular}
\end{table}

\subsection{Comparisons with Other Methods}
In this section, we compare our method with other proposed one-stage detectors on the COCO \emph{test-dev} set. We use the same model and hyperparameters as in previous sections. Follow the convention in RetinaNet, we extend the training schedule to $1.5\times$ and adopt scale jittering from 640 to 800 during training. Compared with other methods, AlignDet achieves much higher AP@0.75, which demonstrates the benefits of aligned features for precise localization.
\begin{table}[hbpt]
\scriptsize
\caption{Comparing AlignDet with other one-stage detectors on the COCO \textit{test-dev} set. M means M40 or Titan X(Maxwell) and P means P100, Titan X(Pascal), Titan Xp or 1080Ti. All speeds on M come from YOLOv3~\cite{yolov3}. $\dagger$ indicates the speed is inferred relatively from RetinaNet. The speed field could only \textbf{remotely} reflect the actual runtime of a model due to the difference in implementations.}
\label{tb:t7-sota}
\renewcommand\arraystretch{1.25}
\begin{threeparttable}
\centering
\begin{tabular}{ccc|cccc|ccc}
	\toprule
	Model                                    & Backbone & Input Size &  Speed/GPU      & AP   & AP$_{50}$ & AP$_{75}$ & AP$_{s}$ & AP$_{m}$ & AP$_{l}$ \\
	\midrule
	YOLOv3~\cite{yolov3}                     & DarkNet-53     & 608 & 51ms/M            & 33.0 & 57.9 & 34.4 & 18.3 & 35.4 & 41.9 \\
	SSD~\cite{ssd,dssd}                      & ResNet-101     & 513 & 125ms/M          & 31.2 & 50.4 & 33.3 & 10.2 & 34.5 & 49.8 \\
	DSSD~\cite{dssd}                         & ResNet-101     & 513 & 156ms/M          & 33.2 & 53.3 & 35.2 & 13.0 & 35.4 & 51.1 \\
    RefineDet~\cite{refinedet}               & ResNet-101     & 512 & -                 & 36.4 & 57.5 & 39.5 & 16.6 & 39.9 & 51.4 \\
    CornetNet~\cite{cornernet}$^{\square\triangle}$   & Hourglass-104  & 511 & 300ms/P          & 40.5 & 56.5 & 43.1 & 19.4 & 42.7 & 53.9 \\
    ExtremeNet~\cite{extremenet}$^{\square\triangle}$ & Hourglass-104  & 511 & 322ms/P          & 40.1 & 55.3 & 43.2 & 20.3 & 43.2 & 53.1 \\
    CenterNet~\cite{centernet}$^{\square\triangle}$   & Hourglass-104  & 511 & 340ms/P          & 44.9 & 62.4 & 48.1 & 25.6 & 47.4 & 57.4 \\
    CenterNet~\cite{centernet2}$^\square$             & Hourglass-104  & 511 & 128ms/P          & 42.1 & 61.1 & 45.9 & 24.1 & 45.5 & 52.8 \\
    \midrule
    RetinaNet~\cite{retinanet}       & ResNet-101 FPN & 800 & 104ms/P          & 39.1 & 59.1 & 42.3 & 21.8 & 42.7 & 50.2 \\
    FoveaBox~\cite{foveabox}         & ResNet-101 FPN & 800 & -                 & 40.6 & 60.1 & 43.5 & 23.3 & \bd{45.2} & \bd{54.5} \\
    FCOS~\cite{fcos}                 & ResNet-101 FPN & 800 & -                 & 41.0 & 60.7 & 44.1 & 24.0 & 44.1 & 51.0 \\
    FSAF~\cite{fsaf}                 & ResNet-101 FPN & 800 & 109ms/P$^\dagger$& 40.9 & 61.5 & 44.0 & 24.0 & 44.2 & 51.3 \\
    RPDet~\cite{rpdet}               & ResNet-101 FPN & 800 & -                 & 41.0 & \bd{62.9} & 44.3 & 23.6 & 44.1 & 51.7 \\
    \bd{AlignDet(Ours)}              & ResNet-101 FPN & 800 & 110ms/P          & \bd{42.0} & 62.4 & \bd{46.5} & \bd{24.6} & 44.8 & 53.3 \\
    \midrule
    RetinaNet~\cite{retinanet}       & ResNeXt-101-32$\times$8d FPN & 800 & 177ms/P           & 40.8 & 61.1 & 44.1 & 24.1 & 44.2 & 51.2 \\
    FoveaBox~\cite{foveabox}         & ResNeXt-101-32$\times$8d FPN & 800 & -                  & 42.1 & 61.9 & 45.2 & 24.9 & 46.8 & \bd{55.6} \\
    FCOS~\cite{fcos}                 & ResNeXt-101-32$\times$8d FPN & 800 & -                  & 42.1 & 62.1 & 45.2 & 25.6 & 44.9 & 52.0 \\
    FSAF~\cite{fsaf}                 & ResNeXt-101-64$\times$4d FPN & 800 & 188ms/P$^\dagger$ & 42.9 & 63.8 & 46.3 & 26.6 & 46.2 & 52.7 \\
    \bd{AlignDet(Ours)}              & ResNeXt-101-32$\times$8d FPN & 800 & 180ms/P           & \bd{44.1} & \bd{64.7} & \bd{48.9} & \bd{26.9} & \bd{47.0} & 54.7 \\
	\bottomrule
\end{tabular}
\begin{tablenotes}
\item $^\square$ indicates using flip test
\item $^\triangle$ indicates using soft NMS
\end{tablenotes}
\end{threeparttable}
\end{table}

\section{Conclusion}
In this work, we investigate the misalignment issue in one-stage detectors. We first discover the close connection between convolution and existing region feature extractors. Based on our findings, we propose a novel RoIConv operator which aligns features with its corresponding bounding box effectively and efficiently for one-stage detectors. Then based on RoIConv, we propose an AlignDet detector, which is fast and performant. Benchmarks on large scale dataset and detailed analyses verify the strength of our proposed method.

\bibliographystyle{plain}
\bibliography{aligndet}

\end{document}